\documentclass[runningheads]{llncs}
\usepackage{algorithm}
\usepackage{algorithmic}
\usepackage{booktabs}
\usepackage{todonotes}
\usepackage{amssymb, amsmath,array, amscd}
\usepackage{bbm}
\usepackage{physics}
\usepackage[T1]{fontenc}
\usepackage{graphicx}
\usepackage{booktabs}
\usepackage{multirow}
\usepackage{relsize}
\usepackage[colorlinks=true, citecolor=black,linkcolor=black,urlcolor=blue]{hyperref}
\usepackage{color}

\makeatletter
\newcommand{\printfnsymbol}[1]{%
  \textsuperscript{\@fnsymbol{#1}}%
}

\begin{document}

\title{Adversarially Robust Prototypical Few-shot Segmentation with Neural-ODEs}
\author{Prashant Pandey\thanks{equal contribution}\inst{1}\orcidID{0000-0002-6594-9685}   \and
Aleti Vardhan\printfnsymbol{1}\inst{2} \and
Mustafa Chasmai\inst{1} \and 
Tanuj Sur\inst{3} \and
Brejesh Lall\inst{1}}
\index{Pandey, Prashant} 
\index{Vardhan, Aleti} 
\index{Chasmai, Mustafa}
\index{Sur, Tanuj} 
\index{Lall, Brejesh} 
\authorrunning{Pandey et al.}
\institute{Indian Institute of Technology Delhi, India \and
Manipal Institute of Technology, India\\
\and
Chennai Mathematical Institute, India\\
\email{getprashant57@gmail.com}
}
\maketitle              % typeset the header of the contribution
\begin{abstract}
% The abstract should briefly summarize the contents of the paper in
% 15--250 words.
Few-shot Learning (FSL) methods are being adopted in settings where data is not abundantly available. This is especially seen in medical domains where the annotations are expensive to obtain. Deep Neural Networks have been shown to be vulnerable to adversarial attacks. This is even more severe in the case of FSL due to the lack of a large number of training examples. In this paper, we provide a framework to make few-shot segmentation models adversarially robust in the medical domain where such attacks can severely impact the decisions made by clinicians who use them.
We propose a novel robust few-shot segmentation framework, Prototypical Neural Ordinary Differential Equation (PNODE), that provides defense against gradient-based adversarial attacks.
We show that our framework is more robust compared to traditional adversarial defense mechanisms such as adversarial training. Adversarial training involves increased training time and shows robustness to limited types of attacks depending on the type of adversarial examples seen during training. Our proposed framework generalises well to common adversarial attacks like FGSM, PGD and SMIA while having the model parameters comparable to the existing few-shot segmentation models. We show the effectiveness of our proposed approach on three publicly available multi-organ segmentation datasets in both in-domain and cross-domain settings by attacking the support and query sets without the need for ad-hoc adversarial training.

\keywords{Few-shot Segmentation\and Neural-ODE\and Adversarial Robustness.}
\end{abstract}

\section{Introduction}
Modern
day safety-critical medical systems are vulnerable to different kinds of attacks that can cause danger to life. With the penetration of AI, Machine Learning and Deep Neural models to healthcare and medical systems, it is imperative to make such models robust against different kinds of attacks. By design, these models are data-hungry and need a significant amount of labelled data to improve their performance and generalizability. Past studies have shown that it is not always feasible to annotate medical data, especially for segmentation problems due to the huge time and specific skills it needs to do so. Lack of well-annotated data, make these models vulnerable to different kind of attacks like adversarial white and black box attacks~\cite{Ian,Kurakin,Madry} on Deep Neural models. ML practitioners employ FSL~\cite{Ravi,One-shot} to learn patterns using well-annotated base classes, finally to transfer the knowledge to scarcely annotated novel classes. This knowledge transfer is severely impacted in the presence of adversarial attacks when \textit{support} and \textit{query} samples from novel classes are injected with adversarial noise \cite{Goldblum}.  

%Existing works propose defence mechanisms~\cite{Kolter,Wong} that cater to specific attacks and their generalizability on different kinds of adversarial attacks in the medical segmentation domain is not fully ascertained. 
Commonly used Adversarial Training  mechanisms~\cite{Ian,Madry,Hongyang} require adversarially perturbed examples shown to the model during training.~\cite{Divide} introduced standard adversarial training (SAT) procedure for semantic segmentation. These methods do not guarantee defense when the type of attack is different from the adversarially perturbed examples~\cite{blind,Park} and it is impractical to expose the model with different kind of adversarial examples during training itself. Also, a common method that handles attacks both on support and query examples of novel classes, is non-existent. To the best of our knowledge, the adversarial attacks on few-shot segmentation (FSS) with Deep Neural models and their defense mechanisms have not yet been explored and the need for such robust models is inevitable. To this end, we propose \textbf{P}rototypical \textbf{N}eural \textbf{O}rdinary \textbf{D}ifferential \textbf{E}quation (PNODE), a novel prototypical few-shot segmentation framework based on Neural-ODEs \cite{chen2018neuralode} that provides defense against different kinds of adversarial attacks in different settings. Owing to the fact that the integral curves of Neural-ODEs are non-intersecting, adversarial perturbations in the input lead to small changes in the output as opposed to existing FSS models where the output is unpredictable.
%The integral curves of Neural-ODEs \cite{chen2018neuralode} have been shown to be non-intersecting, which ensures that adversarial perturbations in the input lead to small changes in the output as opposed to existing FSS models where the output is unpredictable.
In this paper, we make the following contributions:

%- We propose a novel adversarial training mechanism for few-shot learning that can work in tandem with PNODE to improve performance and generalizability against adversarial attacks.

% - We experiment by attacking both support and query, and extend the standard adversarial training procedure to handle both attacks simultaneously.

- We extend SAT
for FSS task to handle attacks on both support and query.

- We propose a novel adversarially robust FSS framework, PNODE, that can handle different kinds of adversarial attacks like FGSM~\cite{Ian}, PGD~\cite{Madry} and SMIA~\cite{Gege} differing in intensity and design, even without an expensive adversarial training procedure. 

- We show the effectiveness of our proposed approach with publicly available multi-organ segmentation datasets like BCV~\cite{BCV}, CT-ORG~\cite{CTORG} and DECATHLON~\cite{DECATHLON} for both in-domain and cross-domain settings on novel classes.

\section{Related Works}
\textbf{Neural ODEs:}
Deep learning models such as ResNets~\cite{resnet} learn a sequence
of transformation by mapping input \textbf{x} to output \textbf{y} by composing a sequence of
transformations to a hidden state. In a
ResNet block, computation of a hidden layer representation
can be expressed using the following transformation: $\mathrm{\textbf{h}(t+1)} = \mathrm{\textbf{h}(t)} + f_{\theta}\mathrm{(\textbf{h}(t), t)}$
%\begin{equation}
%    \mathrm{\textbf{h}(t+1)} = \mathrm{\textbf{h}(t)} + f_{\theta}\mathrm{(\textbf{h}(t), t)}
 %   \label{eq:resnet}
%\end{equation}
where $\mathrm{t \in \{0, \hdots, T\}}$ and $\mathrm{\textbf{h}}:[0,\infty] \rightarrow \mathbb{R}^n$. 
As the number of layers are increased and smaller
steps are taken, in the limit, the continuous
dynamics of the hidden layers are parameterized using an ordinary differential
equation (ODE)~\cite{chen2018neuralode} specified by a neural network $\frac{d\textbf{h}\mathrm{(t)}}{d\mathrm{t}} = f_{\theta}\mathrm{(\textbf{h}(t), t)}$
%\begin{equation} \label{eq:2}
%    \frac{d\textbf{h}\mathrm{(t)}}{d\mathrm{t}} = f_{\theta}\mathrm{(\textbf{h}(t), t)}
%\end{equation}
where $f:\mathbb{R}^n \times [0,\infty] \rightarrow \mathbb{R}^n$
denotes the non-linear trainable layers parameterized by
weights $\mathbf{\theta}$ and \textbf{h} represents the $n$-dimensional state of the Neural-ODE.
These layers define the relation between the input $\mathrm{h(0)}$ and output $\mathrm{h(T)}$, at time $\mathrm{T}>0$,  by providing solution to the ODE initial value problem at terminal time T. Neural-ODEs
are the continuous equivalent of ResNets where the hidden layers can
be regarded as discrete-time difference equations.\\
%as shown in equation~\ref{eq:resnet}.\\
Recent studies~\cite{Yan,Liu,Kang} have applied Neural-ODEs to defend
against adversarial attacks.~\cite{Yan} proposes time-invariant steady Neural-ODE that is more stable than conventional convolutional neural networks (CNNs) in the classification setting. \\
% To design adversarially robust prototypical FSS framework, we leverage the property that the integral
% curves from an ODE solution, starting from different initial points, do not intersect and
% preserve uniqueness in the solution function space.\\
%~\cite{Kang} design a neural ODE such that the features extracted are within a
%neighborhood of the Lyapunov-stable equilibrium points of the ODE. They propose a stable neural ODE for defending against adversarial attacks (SODEF) to
%suppress the input perturbations. We are first to use the inherent robustness of ODEs to make few-shot segmentation model more robust.  \\ 
%However, neither the non-intersecting property nor the steady-state constraint
%used in TisODE can guarantee robustness against input perturbations since these constraints do not
%ensure that the inputs are within a neighborhood of Lyapunov-stable equilibrium points.
\textbf{Few-shot Learning:} 
FSL methods seek good generalization and learn transferable knowledge across different tasks with limited data~\cite{One-shot,Zhao,Cheng}. 
%, typically containing
%just a few training samples of the target classes
Few-shot segmentation (FSS)~\cite{SENet,PANet,FSS} aims to perform pixel-level classification for novel classes in a query image when trained on only a few labelled support images.% ~\cite{Shaban} proposed a novel technique for one-shot segmentation.
The commonly adopted approach for FSS is based on prototypical networks~\cite{snell,PANet,Tang} that employ prototypes to represent typical information for foreground objects present in the support images. 
%The prototype is subsequently compared with the query images that require segmentation via cosine similarity. 
%Deep learning has become an integral part of the medical imaging community
%for a varied range of tasks like classification, segmentation, detection, etc. Especially due to lack of labelled data, application of few-shot learning has gained a lot of momentum in this domain.
%In few-shot medical image segmentation, most works generate new training data to increase the size of the training set given only a few labels~\cite{Zhao,Cheng}. 
%However, whenever a new class needs to be segmented, we need to retrain the model which makes these models difficult to generalize to unseen classes.
In addition to prototype-based setting, \cite{SENet} incorporates `squeeze \& excite’ blocks that avoids the need of pre-trained models for medical image segmentation. \cite{FSS} uses a relation network~\cite{relationNet} and introduced FSS-1000 dataset that is significantly smaller as compared to contemporary large-scale datasets for FSS.
% More recently, few works focus on designing
% network architecture that does not require retraining the
% model. Squeeze and excite~\cite{SENet} proposes a few-shot
% learning architecture specifically designed for medical image
% segmentation. They propose to use squeeze and excite
% modules to fuse information from support image on to query
% image to guide the segmentation arm.~\cite{Tang} propose a new framework for
% few-shot medical image segmentation based on prototypical
% networks. 
\\
%They use a context relation encoder and a recurrent mask refinement module to refine the segmentation mask iteratively.\\
\textbf{Adversarial robustness:}
Adversarial attacks for natural image classification has been extensively explored. FGSM~\cite{Ian} and PGD~\cite{Madry} generate adversarial examples based on the CNN gradients. Besides image classification, several attack
methods have also been proposed for semantic segmentation~\cite{Moosavi,Xie,Gege,Utku}.
% object detection and object tracking~\cite{jia}. In~\cite{Fischer,Dong2}, the classification based attacks were shown
% transferable to attack deep image segmentation results. 
% The universal perturbations were demonstrated 
% existing in~\cite{Moosavi}.
~\cite{Xie} introduced Dense Adversary Generation (DAG) that optimizes a loss function over a set of pixels for generating adversarial perturbations.
% method for both semantic segmentation and object detection attacks. 
% They were first
% to make adversarial examples for semantic segmentation, which is directly related to
% medical image segmentation.
\cite{Paschali} studied the effects of adversarial attacks on brain segmentation and
skin lesion classification. 
%They showed that state-of-the-art networks such as~\cite{Inception} and~\cite{Unet} are
%still extremely susceptible to adversarial examples for skin lesion and brain segmentation. 
%The general idea of natural image attacks was to iteratively generate perturbations based on the CNN
%gradients to maximize the network predictions of adversarial examples and the ground truth labels.
%~\cite{Finlayson} used PGD white and black box attacks on fundoscopy, dermoscopy, and chest X-ray images, using a pre-trained ResNet50 model. %By producing crafted mask, an adaptive segmentation mask attack (ASMA) is proposed to fool DNN model~\cite{Utku}. 
Recently, \cite{Gege} proposes an adversarial attack (SMIA) for images in medical domain that employs a loss stabilization term to exhaustively search the perturbation space.
% to consistently produce adversarial perturbations on medical images. 
While adversarial attacks expose the vulnerability of deep neural networks, adversarial training~\cite{Madry,Ian,Alexey} is effective in enhancing the target model by training it with adversarial samples.
%~\cite{Divide} introduced a standard adversarial training procedure for semantic segmentation.
However, none of existing methods have explored SAT procedure for few-shot semantic segmentation.

\begin{figure}[t!]
\centering
\includegraphics[width=4.20in, height=2.24in]{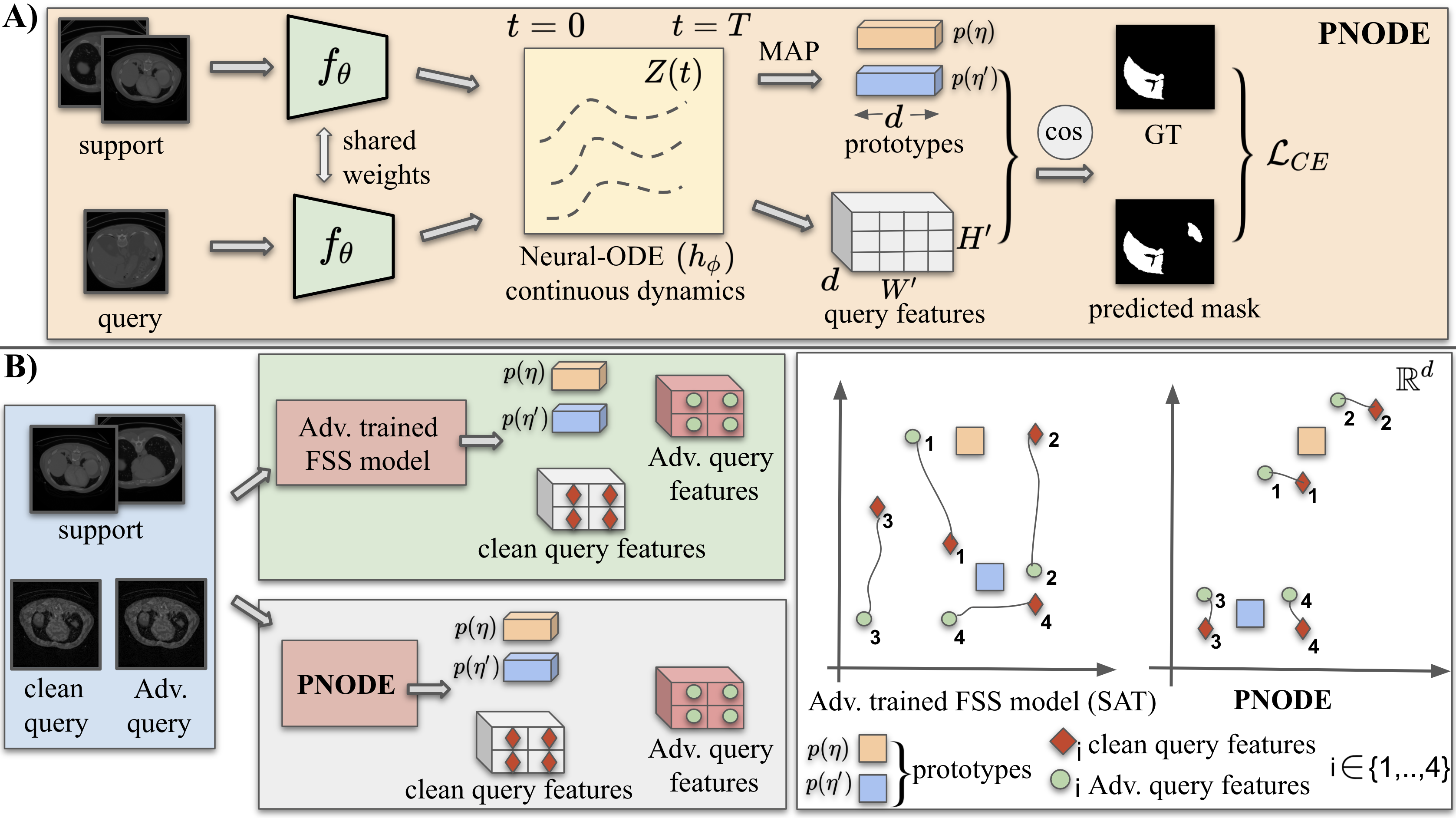}
% \caption{\textbf{A)} The feature extractor $f_\theta$ produces intermediate representations $Z^{k}_{S}$ and query $Z_{Q}$ for support and query images. $t=0$ for the Neural ODE. The final output features $Z_{S}(T)$ and $Z_{Q}(T)$ at time T. These output features depend on the initial states $Z_{S}(0)$,$Z_{Q}(0)$  and the dynamics of the Neural-ODE denoted by $h_{\phi}$ (weights of the neural ODE layers). Prototypes for foreground class $\eta \in \mathrm{C}$ and background class $\eta' $ are obtained by applying Masked Average Pool on the support features $Z_{S}(T)$ and their corresponding label masks. The query mask prediction is done by computing the cosine similarities between all the pixel features in $Z_{Q}(T)$ and the obtained prototypes. \textbf{B)} The working of PNODE is demonstrated by visualizing the d-dimensional representation space to which the clean and adversarial query features are mapped. Due to the non-intersecting property of integral curves in neural ODEs, intrinsic constraints are imposed on the features which make adv. query feature outputs move closer the the clean query feature outputs unlike traditional FSS models. Further, we highlight the limitations of traditional defense methods (Standard Adv. Training) by attacking the adversarially trained models with attacks they aren't trained on to show that PNODE is better at handling various unseen attacks. Similarly, these attacks can be extended to support as well.  
% }
\caption{\textbf{A)} Robust features for query and support images are obtained by the feature extractor followed by the continuous dynamics and integral solutions of a Neural-ODE. Class-wise prototypes are obtained by applying Masked Average Pool (MAP) on support features. Pixel-level cosine similarities of query features with the prototypes provide query mask predictions. \textbf{B)} $d$-dimensional representation of clean and adversarial query features. Adversarial query features lie closer to the clean ones unlike Adversarially trained FSS model (or SAT). In SAT, perturbations from one class  may be closer to prototypes of another class (model is confused), while  for PNODE, they tend to remain closer. 
}
\label{fig:pnode_fig}
\end{figure}
\section{Proposed Method}
The objective is to build a FSS model robust to various gradient-based attacks on support and query images. Our methodology focuses on two aspects. First, we extend SAT as a defense mechanism. Second, we propose our framework, PNODE, which alleviates existing limitations faced by SAT.

\subsection{Problem setting}
 %The objective is to train a robust few shot segmentation model on the training set $\mathcal{D_{\mathrm{train}}}$ and evaluate it on test set $\mathcal{D_{\mathrm{test}}}$. $\mathcal{D_{\mathrm{train}}}$ and $\mathcal{D_{\mathrm{test}}}$ are constructed using class sets $C_{\mathrm{train}}$ and $C_{\mathrm{test}}$ respectively.
FSS setting includes train $\mathcal{D}_{\mathrm{train}}$ and test $\mathcal{D}_{\mathrm{test}}$ datasets having non-overlapping class sets. 
% ($C_{\mathrm{train}} \cap C_{\mathrm{test}} = \phi$)
% $C_{\mathrm{train}}$ and $\mathcal{C_{\mathrm{test}}}$ are mutually exclusive i.e. $C_{\mathrm{train}} \cap C_{\mathrm{test}} = \phi $.
% The train and test datasets are formed using 
% Both the training $\mathcal{D_{\mathrm{train}}}$ set and the test $\mathcal{D_{\mathrm{test}}}$ set consist
% of several episodes such that each episode comprises of a support set $\mathcal{S}$
% % with its
% % corresponding binary annotation map $L_{S}(\eta)$ with respect to
% % the semantic class (or organ) $\eta$ 
% and a query set $\mathcal{Q}$.
 Each dataset consists of a set of episodes with each episode containing a $N$-way $K$-shot task $\mathcal{T}_{i}$ =  ($\mathcal{S}_{i}, \mathcal{Q}_{i}$) where $\mathcal{S}_i$ and $\mathcal{Q}_i$ are support and query sets for the $i^{th}$ episode having class set $C_{i}$.
Formally, $\mathcal{D_{\mathrm{train}}} = \{(\mathcal{S}_i,\mathcal{Q}_i)\}^{E_{\mathrm{train}}}_{i = 1}$ and $\mathcal{D_{\mathrm{test}}} = \{(\mathcal{S}_i,\mathcal{Q}_i)\}^{E_{\mathrm{test}}}_{i = 1}$ where $E_{\mathrm{train}}$ and $E_{\mathrm{test}}$ denote the number of episodes during training and testing.
%The semantic classes for training and testing
% are mutually exclusive i.e. $C_{\mathrm{train}} \cap C_{\mathrm{test}} = \phi $.\\
The support set $\mathcal{S}_i$ has $K$ image $(\mathcal{I}_{\mathcal{S}})$, mask $(L_{\mathcal{S}})$ pairs per class
with a total of $N$ semantic classes  i.e. $\mathcal{S}_i =
\{(\mathcal{I}_{\mathcal{S}}^{k},L_{\mathcal{S}}^{k}(\eta))\}$ where  $L^{k}_{\mathcal{S}}(\eta)$ is the ground-truth mask for $k$-th shot corresponding to class$~\eta \in C_{i}$, $|C_{i}| = N$ and $k=1,2, \hdots, K$.
%  Hence for any episode, the support set $\mathcal{S}_i $ contains $N \times K$ samples 
The query set $\mathcal{Q}_i$ has $N_{\mathcal{Q}}$ $\mathcal{\mathrm{image}~(I_{Q})},
{\mathrm{mask}~(L_{\mathcal{Q}})}$ pairs.
%The objective is to learn a semantic segmentation network %$\mathcal{F(.)}$ from
%$\mathcal{D_{\mathrm{train}}}$, such that given a support set %$\mathcal{S}_{\mathrm{novel}} = \{(\mathcal{I}_{S},L_{S}(\eta'))\} \notin \mathcal{D_{\mathrm{train}}}$
%for a novel semantic class $\eta' \in C_{\mathrm{novel}}$ and a query image $\mathcal{I}_Q$,
%we can predict the binary segmentation $M_Q(\eta')$ of the query.\\
%ction{Adversarial Attacks}
% Given an episode $i$ with few-shot semantic segmentation network $\mathcal{F(.)}$, support set ($\mathcal{S}_{i}$) and query image-mask pair ($\mathcal{I}_{Q}$,$L_{Q}$$(\eta)$), the output is a segmentation mask ${L'_{Q}} = \mathcal{F(S, I_Q)}$, where
% $\mathcal{I}_{S}^{k},\mathcal{I_{Q}}\in\mathbb{R}^{H \times W \times 1}$ and ${L_Q} \in \mathbb{R}^{H \times W\times (N+1)}$ - $H,W$ and $N$ are the height,
% width and number of classes in $\mathcal{I}_{Q}$ respectively. \\
The FSS model $\mathcal{F}(.)$ is trained on $\mathcal{D}_{\mathrm{train}}$ across the episodes with support sets and query images as inputs, and predicts the segmentation mask $M_{\mathcal{Q}}$ = $\mathcal{F}(\mathcal{S}_i, \mathcal{I}_{\mathcal{Q}})$ in the $i$-th episode for query image $\mathcal{I}_\mathcal{Q}$. During testing, the trained model $\mathcal{F}(.)$ is used to predict masks for unseen novel classes with the corresponding support set samples and query images as inputs from $\mathcal{D}_{\mathrm{test}}$.  \\
 Further, the trained FSS model is adversarially attacked to record the drop in performance. An adversarial version of a clean sample can be generated by exploiting gradient information from the model  $\mathcal{F}(.)$ employing \cite{Ian}. Specific to the case of FSS, the prediction of query mask not only depends on the query image but also on the information from support set. This enables the attacks to be designed in such a way that either attacked query or support can deteriorate the query prediction. These perturbations are specifically chosen so that the loss between ground-truth and the predicted masks of the query increases. 

\subsection{Adversarial Training}

%To build a defense mechanism, we extend SAT procedure
%to the FSS task as shown in \ref{alg:MYALG}. 
To ensure that the segmentation model $\mathcal{F}(.)$ is robust to adversarial perturbations for both support and query images, we extend $\mathcal{D_{\mathrm{train}}}$ in each batch during training with two additional batches generated for the $i^{th}$ episode using update rule from~\cite{Ian} as follows: (a) 
%  Given $\langle$image, mask$\rangle$ pair from support set $\mathcal{S}_{i}$ 
% and query set $\{(\mathcal{I}_{\mathcal{Q}},L_{\mathcal{Q}_{i}}(\eta))\}$ from $\mathcal{Q}_{i}$,
generate adversarial example for support image $\mathcal{I}_{\mathcal{S}}$:
% using the FGSM update rule~\cite{Ian}:
\begin{equation}
    \mathcal{I^{\mathrm{adv}}_S} = \mathcal{I_S} + \epsilon.\mathrm{sign}(\nabla_{\mathcal{I_S}}\mathcal{L(F}(\mathcal{S}_i, \mathcal{I_Q}), L_{\mathcal{Q}}(\eta)))\label{eq:FGSM_FSS1}
\end{equation}
%\begin{equation}
%    \mathcal{I^{\mathrm{adv}}_S} = \mathcal{I^{\mathrm{orig}}_S} + \epsilon.\mathrm{sign}(\nabla_{\mathcal{I^{\mathrm{orig}}_S}}\mathcal{L(F}(\mathcal{I^{\mathrm{orig}}_S}, \mathcal{I^{\mathrm{orig}}_Q}), L(\eta)))\label{eq:FGSM_FSS1}
%\end{equation}
%On applying this to $\mathcal{D^{\mathrm{orig}}_{\mathrm{train}}}$, we create $\mathcal{D^{\mathcal{S}}_{\mathrm{train}}}$.\\
(b) generate adversarial example for query image $\mathcal{I}_{\mathcal{Q}}$:
\begin{equation}
    \mathcal{I^{\mathrm{adv}}_Q} = \mathcal{I_Q} + \epsilon.\mathrm{sign}(\nabla_{\mathcal{I_Q}}\mathcal{L(F}(\mathcal{S}_i, \mathcal{I_Q}), L_{\mathcal{Q}}(\eta)))\label{eq:FGSM_FSS2}
\end{equation}
This is a single-step attack, which minimises the $l_{\infty}$ norm of the perturbation bounded by parameter $\epsilon$.
% \begin{equation}
%     \mathcal{I^{\mathrm{adv}}_Q} = \mathcal{I_Q} + %\epsilon.\mathrm{sign}(\nabla_{\mathcal{I_Q}}\mathcal{L(F}(\mathcal{I_S}%, \mathcal{I_Q}), L(\eta)))\label{eq:FGSM_FSS2}
% \end{equation}
% On applying this to $\mathcal{D^{\mathrm{orig}}_{\mathrm{train}}}$, we create $\mathcal{D^{\mathcal{Q}}_{\mathrm{train}}}$.\\
The detailed procedure of SAT is listed in Algorithm~\ref{alg:SAT}.
%In each iteration of SAT, we first calculate the loss $\mathcal{L}$ based on $\mathcal{D^{\mathrm{orig}}_{\mathrm{train}}}$. Using the gradients $\nabla_{\mathcal{I^{\mathrm{orig}}_S}}\mathcal{L}, \nabla_{\mathcal{I^{\mathrm{orig}}_Q}}\mathcal{L}$, we generate $\mathcal{D^{\mathrm{adv}}_{\mathrm{train}}} = \{\mathcal{D^{\mathcal{S}}_{\mathrm{train}}}, \mathcal{D^{\mathcal{Q}}_{\mathrm{train}}}\}$ based on methods (1) and (2).
%$\mathcal{F}$ is further trained on $\mathcal{D^{\mathrm{adv}}_{\mathrm{train}}}$ and the model parameters are updated accordingly.

\begin{algorithm}
\caption{Standard Adversarial Training (SAT) for FSS}
\begin{algorithmic}[1] 
% \algsetup{linenosize=\tiny}
%  \scriptsize
\REQUIRE Clean training data $\mathcal{D_{\mathrm{train}}} = \{(\mathcal{S}_i,\mathcal{Q}_i)\}^{E_{train}}_{i = 1}$, segmentation network $\mathcal{F}(.)$.
% , number of training iterations $\tau$.
% \STATE Set i = 1.%, $b = $ batch size.
\FOR{$\mathrm{i} \in \{1,\hdots, E_{train}\}$}
%\STATE Train $\mathcal{F}$ on $\mathcal{D^{\mathrm{orig}}_{\mathrm{train}}}$. Calculate gradients w.r.t support and query images.
\STATE Sample episode $\mathcal{E}^{\mathrm{orig}}=\{(\mathcal{S}_{\mathrm{i}},\mathcal{Q}_{\mathrm{i}})\}$ from $\mathcal{D_{\mathrm{train}}}$.
\STATE Calculate gradients w.r.t support and query images using $\mathcal{F}(.)$.
\STATE Get  $\mathcal{\mathcal{E}^{\mathcal{S}}} = \{(\mathcal{S}^{\mathrm{adv}}_i,\mathcal{Q}_i)\}$ by perturbing support images using equation~\ref{eq:FGSM_FSS1}.
\STATE Get $\mathcal{\mathcal{E}^{\mathrm{Q}}} = \{(\mathcal{S}_i, \mathcal{Q}^{\mathrm{adv}}_i)\}$ by perturbing query images using equation~\ref{eq:FGSM_FSS2}.
%\STATE Combine data $\mathcal{D^{\mathrm{adv}}_{\mathrm{train}}} = \{\mathcal{D^{\mathcal{S}}_{\mathrm{train}}}, \mathcal{D^{\mathcal{Q}}_{\mathrm{train}}}\}$ with $\mathcal{D^{\mathrm{orig}}_{\mathrm{train}}}$.
\FOR{$ 
\mathcal{E} \in \{\mathcal{E^{\mathrm{orig}}}, \mathcal{E^{\mathcal{S}}}, \mathcal{E^{\mathcal{Q}}}\}$}
\STATE Train $\mathcal{F}$ on  episode $\mathcal{E}$.
\ENDFOR
% \STATE $i\leftarrow i + 1$

% for D in {D_train, D_train_S, D_train_Q}:
%     Train F on D
%     compute loss update params
%\STATE Set batch as $\{(\mathcal{S}^{\mathrm{orig}}_1,\mathcal{Q}^{\mathrm{orig}}_1), \hdots, (\mathcal{S}^{\mathrm{orig}}_{\mathrm{N_{train}}},\mathcal{Q}^{\mathrm{orig}}_{\mathrm{N_{train}}}),  (\mathcal{S}^{\mathrm{adv}}_1,\mathcal{Q}^{\mathrm{orig}}_1), \hdots, (\mathcal{S}^{\mathrm{orig}}_{\mathrm{N_{train}}},\mathcal{Q}^{\mathrm{adv}}_{\mathrm{N_{train}}})\}$
%\STATE Train $\mathcal{F}$ on %$\mathcal{D^{\mathrm{adv}}_{\mathrm{train}}}$. Compute the loss. Update parameters of $\mathcal{F}$. $i\leftarrow i + 1$
\ENDFOR
\end{algorithmic}
\label{alg:SAT}
\end{algorithm}

SAT requires prior knowledge on the type of adversarial attacks to include the corresponding samples during training which is practically infeasible, compute intensive and also doesn't guarantee robustness to unseen attacks. Our PNODE framework addresses these shortcomings.

\subsection{Prototypical Neural ODE (PNODE)}

%   The proposed model learns representations which are robust to adversarial attacks which comes from robust representations for both support and query images. These robust representations are then used to predict the mask for the query image.
The proposed framework is based on existing prototypical few-shot segmentation models \cite{Dong,PANet}. 
Given an episode $i$ with task $\mathcal{T}_{i}$ =  ($\mathcal{S}_{i}, \mathcal{Q}_{i}$),
the feature extractor $f_{\theta}$ generates intermediate feature representations ${Z}^{k}_\mathcal{S}$ and ${Z}_\mathcal{Q}$ for the support and query images $\mathcal{I}^{k}_{\mathcal{S}}$,  $\mathcal{I}_{\mathcal{Q}}$.  
% The Neural-ODE block considers features as a function of time. 
The outputs from the feature extractor $f_{\theta}$ are considered as initial states for the Neural-ODE block at time $t$=0, denoted as ${Z}^{k}_{\mathcal{S}}(0)$, ${Z}_{\mathcal{Q}}(0)$. 
% Outputs from the feature extractor $f_{\theta}$ are considered as the initial states for the solution of the Neual-ODE.

The Neural-ODE block consists of hidden layers $h_{\phi}$ parameterized by $\phi$ and its dynamics are governed by $h_{\phi}$ which control how the intermediate state changes at any given time $t$. The output representation at fixed terminal time $T (T>0)$ for query features $Z_\mathcal{Q}$ is given by $Z_{\mathcal{Q}}(T) = Z_{\mathcal{Q}}(0) +\int ^{T}_{0}h_{\phi }(Z_{\mathcal{Q}}(t),t)dt$.
%\begin{equation} \label{eq:ODE}
%\begin{split}
%    Z_{\mathcal{Q}}(T) = Z_{\mathcal{Q}}(0) +\int ^{T}_{0}h_{\phi %}(Z_{\mathcal{Q}}(t),t)dt  \quad  \mathrm{where} \quad  %\frac{d{Z_{\mathcal{Q}}}(t)}{d\mathrm{t}} = h_{\phi}({Z}_{\mathcal{Q}}(t), %t)
%\end{split}
%\end{equation}
Similarly, the output representation at fixed terminal time $T (T>0)$ for support features $Z^{k}_{\mathcal{S}}$ are generated.
The support feature maps $Z^{k}_{\mathcal{S}}(T)$ from the Neural-ODE block of spatial dimensions ($H' \times W'$) are upsampled to the same spatial dimensions of their corresponding masks ${L}_{\mathcal{S}}$ of dimension $(H\times W$). 
% In the literature the prototypes are either formed using Global Average Pooling or Masked Average Pooling (MAP). 
Inspired by late fusion~\cite{PANet} where the ground-truth labels are masked over feature maps, we employ Masked Average Pooling (MAP) between
$Z^{k}_{\mathcal{S}}(T)$ and ${L}^{k}_{\mathcal{S}}(\eta)$ to form a $d$-dimensional prototype $p(\eta)$ for each foreground class $\eta \in C_{i}$ as shown: 
% The prototype mean among all the K examples is considered as the final prototype.
\begin{equation}\label{eq:MAP}
    p(\eta)=\dfrac{1}{K}\mathlarger{\sum_{k}}\dfrac{\sum _{x,y}\{Z^{k}_{\mathcal{S}}(T)\}^{(x,y)}\cdot \mathbbm{1}{[ \{L^{k}_{\mathcal{S}}(\eta)\}^{(x,y)}=\eta]}}{\sum _{x,y}\mathbbm{1}{[ \{L^{k}_{\mathcal{S}}(\eta)\}^{(x,y)}=\eta]}}
\end{equation}
where $(x,y)$ are the spatial locations in the feature map and $\mathbbm{1}(.)$ is an indicator function. 
The background is also treated as a separate class and the prototype for it is calculated by computing the feature mean of all the spatial locations excluding the ones that belong to the foreground classes.
% Given a support set $\mathcal{S}_i = \{(\mathcal{I}_{S}^{k},L^{k}_{S}(\eta))\}$, the feature map output of $h_{\phi}(f_{\theta}())$ for $\mathcal{I}_{S}^{k}$ is $Z^{k}_{\mathcal{S}}$, where $\eta$ $\in C$  and $k=1,2, \hdots, K$.
% In \eqref{eq:MAP}, $(x,y)$ are the spatial locations in the feature map and $\mathbbm{1}$ is an indicator function. 
\\
The probability map over semantic classes $\eta$ is computed by measuring the cosine similarity (cos) between each of the spatial locations in $Z_{\mathcal{Q}}(T)$
%\todo{replaces query word in suffix with Q} 
with each prototype $p(\eta)$ as given by:
%\todo{Use sumbol used in eq 7} 
\begin{equation}
{{M}}_{\mathcal{Q}}^{(x,y) }(\eta) = \dfrac{\mathrm{exp}(\mathrm{cos}(\{Z_{\mathcal{Q}}(T)\}^{(x,y)},p(\eta)) )}{\sum _{\eta'\in \mathcal{C}_{i}}\mathrm{exp}{(\mathrm{cos}( \{Z_{\mathcal{Q}}(T)\}^{(x,y)},p(\eta')) }}
\end{equation}
The predicted mask $M_{\mathcal{Q}}$ is generated by taking the argmax of $M_{\mathcal{Q}}(\eta)$ across semantic classes. We use Binary Cross Entropy loss $\mathcal{L}_{CE}$ between $M_{\mathcal{Q}}$ and the ground-truth mask for training. For detailed overview of PNODE, refer to Fig.\ref{fig:pnode_fig}.
% The mask of the query is computed by measuring the cosine similarity (cos) between each of the spatial locations in $Z_{\mathcal{Q}}(T)$
% %\todo{replaces query word in suffix with Q} 
% with each prototype $p(\eta)$.
% %\todo{Use sumbol used in eq 7} 
% Following this, softmax is computed for the similarity measures at each spatial location. Each spatial location is classified with class of prototype that it is closest to and this is done by taking the argmax of the softmax output.The final mask is generated by upsampling using bilinear interpolation to match the same spatial dimensions as the ground-truth query mask.
% \begin{equation}
% {{M}}_{\mathcal{Q}}^{(x,y) }(\eta) = \dfrac{\mathrm{exp}(\mathrm{cos}(\{Z_{\mathcal{Q}}(T)\}^{(x,y)},p(\eta)) )}{\sum _{\eta'\in \mathcal{C}_{i}}\mathrm{exp}{(\mathrm{cos}( \{Z_{\mathcal{Q}}(T)\}^{(x,y)},p(\eta')) }}
% \end{equation}
% Finally Binary Cross Entropy $\mathcal{L}_{CE}$ loss is used between the predicted query mask and the true label to train the model. 
\\
% During evaluation, PNODE is robust against adversarial attacks attributed to the fact that the non-intersecting property of integral curves in Neural-ODEs indicated that features starting from some point are constrained by the integral curves starting from that feature's neighbourhood. Thus, in PNODE, if a correctly classified image is adversarially perturbed, the integral curve associated to its perturbed version compared to the original one. Consequently, the perturbed sample's feature representations are close to the originial feature representations as shown in Fig~\ref{fig:pnode_fig} which leads to accurate predictions of the query masks.
% During evaluation, PNODE is robust against adversarial attacks attributed to the fact that the features are constrained by the non-intersecting integral curves.
During evaluation, PNODE's robustness against adversarial attacks is attributed to
the fact that the integral curves corresponding to the features are non-intersecting.
Thus, if a clean sample (support or query) is adversarially perturbed, the integral curves associated with other similar samples constrain the adversrial features to remain bounded in the representation space. Consequently, the perturbed sample's feature representations are closer to the clean or original feature representations as shown in Fig.\ref{fig:pnode_fig} which leads to accurate predictions of the query masks.

\section{Implementation Details} 

%Our framework consists of a CNN-based feature extractor followed by a Neural ODE block consisting of 3 convolutional layers. The solution for Neural ODE is obtained by using the ODE solver Runge-Kutta of order 5 of Dormand-Prince-Shampine.
% The extracted features are then passed through a continuous Neural ODE solver consisting of 3 convolutional layers as its hidden unit dynamics function.
%The ODE  solver evaluates from $t=0$ to $t=4$, and the value of the solution at $t=4$ is used for further non-parametric prototypical learning. Our implementation is built on top of \emph{torchdiffeq} \cite{torchdiffeq}, an opensource library of differentiable ODE solvers.
PNODE framework consists of a CNN-based feature extractor followed by a Neural-ODE block consisting of 3 convolutional layers. The architecture of PNODE consists of a total of 14.7M trainable parameters, while PANet and FSS1000 have 14.7M and 18.6M parameters, respectively. The solution for Neural-ODE is obtained by using Runge-Kutta ODE solver~\cite{chen2018neuralode}. 
% The extracted features are then passed through a continuous Neural ODE solver consisting of 3 convolutional layers as its hidden unit dynamics function.
%The ODE  solver evaluates from $t=0$ to $t=4$, and the value of the solution at $t=4$ is used for further non-parametric prototypical learning. Our implementation is built on top of \emph{torchdiffeq} \cite{torchdiffeq}, an opensource library of differentiable ODE solvers.
% \subsubsection{Adversarial Training}
To understand the effect of adversarial training on prototype-based networks, we employ SAT with PANet~\cite{PANet} and name it AT-PANET. 
% We experiment with adversarial training on PANet as our baseline is prototypical network.
It is trained with FGSM with $\epsilon = 0.025$. To test the trained models, we perturb the support and query images by setting $\epsilon = 0.02, 0.01~\mathrm{and}~0.04$ for FGSM, PGD and SMIA, respectively. For the iterative adversarial attacks SMIA and PGD, we take 10 iterations each. These hyperparameters for the attacks were chosen so as to keep the perturbed images human perceptible. We use one A100 GPU to conduct our experiments. For statistical significance, we run each experiment twice and report the mean and standard deviation. All our implementations can be found \href{https://github.com/prinshul/Prototype\_NeuralODE\_Adv\_Attack}{here.\footnote{
https://github.com/prinshul/Prototype\_NeuralODE\_Adv\_Attack}}

\section{Experiments and Results}

We experiment on three publicly  available multi-organ segmentation datasets, BCV~\cite{BCV}, CT-ORG~\cite{CTORG}, and Decathlon~\cite{DECATHLON} to evaluate the generalisability of our method. We train on the smaller BCV dataset and use CT-ORG and Decathlon for cross-domain FSS. To have a more uniform size of the test set, we sample 500 random slices per organ from the much larger CT-ORG and Decathlon datasets. For the 3D volumes in all three datasets,
% we extract slices having valid masks and divide them to fixed train, test and validation splits. We do not crop the slices class wise and handle multiple organs in the same slice, since cropping in the test set would require labels and this would lead to an unfair evaluation on the test set. 
we extract slices with valid masks and divide them into fixed train, test, and validation splits. We do not crop the slices class-wise and handle multiple organs in the same slice since cropping in the test set would require labels, leading to an unfair test set evaluation.
For baseline models we use PANet \cite{PANet}, FSS1000 \cite{FSS}, SENet \cite{SENet} and AT-PANET. Of the organs available in these datasets, we report results on Liver and Spleen (as novel classes) due to their medical significance and availability in multiple datasets. 

\begin{figure}
    \centering
    \includegraphics[width=0.73\linewidth]{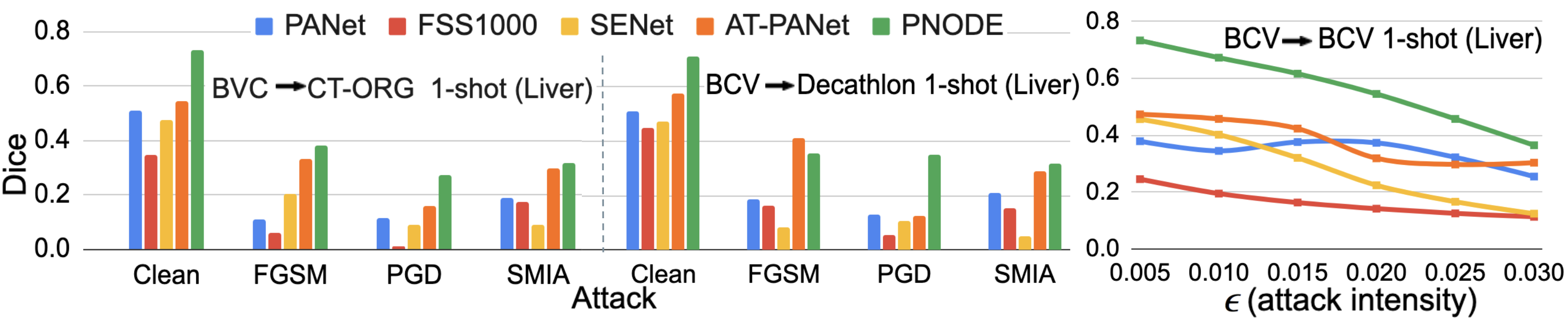}

    \caption{Performance of models for attacks on BCV $\rightarrow$ CT-ORG (left), BCV $\rightarrow$ Decathlon (middle) and for different intensities of FGSM (right), on Liver 1-shot.}
    \label{fig:plots}
\end{figure}

\begin{table}[!ht]
    \centering
    \setlength{\tabcolsep}{2.8pt}
    \renewcommand{\arraystretch}{1.2}
    \caption{1-shot query attack results for BCV $\rightarrow$ BCV in-domain Liver and Spleen organs (novel classes). The dice scores are rounded off to two decimals.}

    \scalebox{0.78}{
    \begin{tabular}{c|cccc|cccc}
    \toprule
        \multirow{2}{*}{Method} & \multicolumn{4}{c|}{BCV $\rightarrow$ BCV (Liver)} & \multicolumn{4}{c}{BCV $\rightarrow$ BCV (Spleen)}\\ 
    % \midrule
         & Clean & FGSM & PGD & SMIA & Clean & FGSM & PGD & SMIA \\
    \midrule
        PANet\cite{PANet} & .61±01 & .29±.01 & .21±.01& .20±.01 & .38±.03 & .16±.01 & .11±.01 & .07±.01   \\
        FSS1000\cite{FSS} & .37±.04 & .10±.03 & .04±.02& .18±.01 & .41±.02 & .19±.01 & .08±.01 & .17±.01   \\
        SENet\cite{SENet} & .61±.01 & .30±.06 & .22±.02& .12±.02& .57±.01 & .04±.01 & .21±.04& .01±.01 \\
      %\midrule
        AT-PANet & .65±.01 & .35±.03 & .27±.02& .36±.01 & .46±.01 & .32±.08 & .19±.03 & .11±.01   \\
        % AT-PNODE & .62 (.00) & \textbf{.38 (.00)} & \textbf{.13 (.00)}& \textbf{.24 (.00)} & .00 (.00) & .00 (.00) & .00 (.00) & .00 (.00)   
        %\\
        \textbf{PNODE} & \textbf{.83±.01} & \textbf{.52±.02} & \textbf{.46±.01}& \textbf{.38±.03} & \textbf{.60±.01} & \textbf{.36±.01} & \textbf{.27±.03} & \textbf{.20±.02}\\
    \bottomrule
    
    \end{tabular}}

    \label{tab:bcv_1shot}
\end{table}
\begin{table}[!ht]
    \centering
    \setlength{\tabcolsep}{2.8pt}
    \renewcommand{\arraystretch}{1.2}
    \caption{3-shot query attack results for Liver BCV in-domain and BCV $\rightarrow$ CT-ORG cross-domain settings. The dice scores are rounded off to two decimals.}

    \scalebox{0.78}{
    \begin{tabular}{c|cccc|cccc}
    \toprule
        \multirow{2}{*}{Method} & \multicolumn{4}{c|}{BCV $\rightarrow$ BCV (Liver)} & \multicolumn{4}{c}{BCV $\rightarrow$ CT-ORG (Liver)}\\ 
    % \midrule
         & Clean & FGSM & PGD & SMIA & Clean & FGSM & PGD & SMIA \\
    \midrule
        PANet\cite{PANet} & .67±.01 & .38±.01 & .31±.01& .16±.01 & .60±.01 & .23±.01 & .21±.01 & .17±.01   \\
        FSS1000\cite{FSS} & .49±.04 & .15±.01 & .05±.01& .19±.03 & .14±.02 & .03±.01 & .01±.01 & .08±.01   \\
        % SENet\cite{SENet} & .61 (.01) & .30 (.06) & .22 (.02)& .12 (.02)& .57 (.00) & .04 (.01) & .21 (.04)& .01 (.01) \\
      %\midrule
        AT-PANet & .69±.01 & .36±.09 & .39±.01& .23±.01 & .64±.03 & .09±.05 & .21±.02 & .18±.01   \\
        % AT-PNODE & .62 (.00) & \textbf{.38 (.00)} & \textbf{.13 (.00)}& \textbf{.24 (.00)} & .00 (.00) & .00 (.00) & .00 (.00) & .00 (.00)   
        %\\
        \textbf{PNODE} & \textbf{.76±.02} & \textbf{.53±.01} & \textbf{.52±.01}& \textbf{.39±.02} & \textbf{.68±.01} & \textbf{.43±.02} & \textbf{.42±.01} & \textbf{.31±.02}\\
    \bottomrule
    
    \end{tabular}}
    \label{tab:3shot}
\end{table}
As shown in Table~\ref{tab:bcv_1shot}, PNODE outperforms all of our baselines by atleast 27\%, 48\%, 70\% and 5\% on clean, FGSM, PGD and SMIA attacks respectively for BCV in-domain Liver setting. While PNODE provides a better defense against the adversarial attacks, it also outperforms the baselines for clean samples. This indicates that PNODE also learns a better representation space of unperturbed support and query samples which is attributed to the continuous dynamics of the Neural-ODE. With small perturbations, the integral curves with respect to the perturbed samples are sandwiched between the curves that correspond to the neighbouring samples ensuring that outputs of the perturbed samples do not change drastically. This is not the case with traditional CNNs, as there are no such intrinsic constraints \cite{Yan}.
To further show the Neural-ODE’s role in robustness, we conduct a set of ablation studies. Upon removing the Neural-ODE block from PNODE, maintaining the remaining architecture and training procedure, we observed 0.41, 0.36, 0.36 and 0.31 units of drop in performance for clean, FGSM, PGD and SMIA, respectively. Further, using SAT made this model more robust, but PNODE outperformed it by 0.28, 0.19, 0.20 and 0.31 units, respectively. An interesting observation is that the baseline results tend to perform well for some attacks, while fail for others. For example, SENet~\cite{SENet} does very well  on PGD~\cite{Madry} attack with a dice of 0.21, but performs very poorly on SMIA~\cite{Gege}. PNODE,  on the other hand performs consistently across the different attacks. As can be seen in Fig.\ref{fig:plots}, PNODE also performs well on a wider range of attack intensities. Some other experimental analyses in the cross-domain setting can be seen in Fig.\ref{fig:plots}, which follow similar patterns with consistently better performance of PNODE. We also perform experiments on the 3-shot setting in Table~\ref{tab:3shot}. While there is a consistent drop between in-domain and cross-domain performance of all models, the drops corresponding to PNODE are relatively smaller. Thus, similar to distribution shifts between clean and perturbed samples, PNODE  is also robust to cross-domain distribution shifts. We visualise
\begin{figure}
    \centering
    \includegraphics[width=0.67\linewidth]{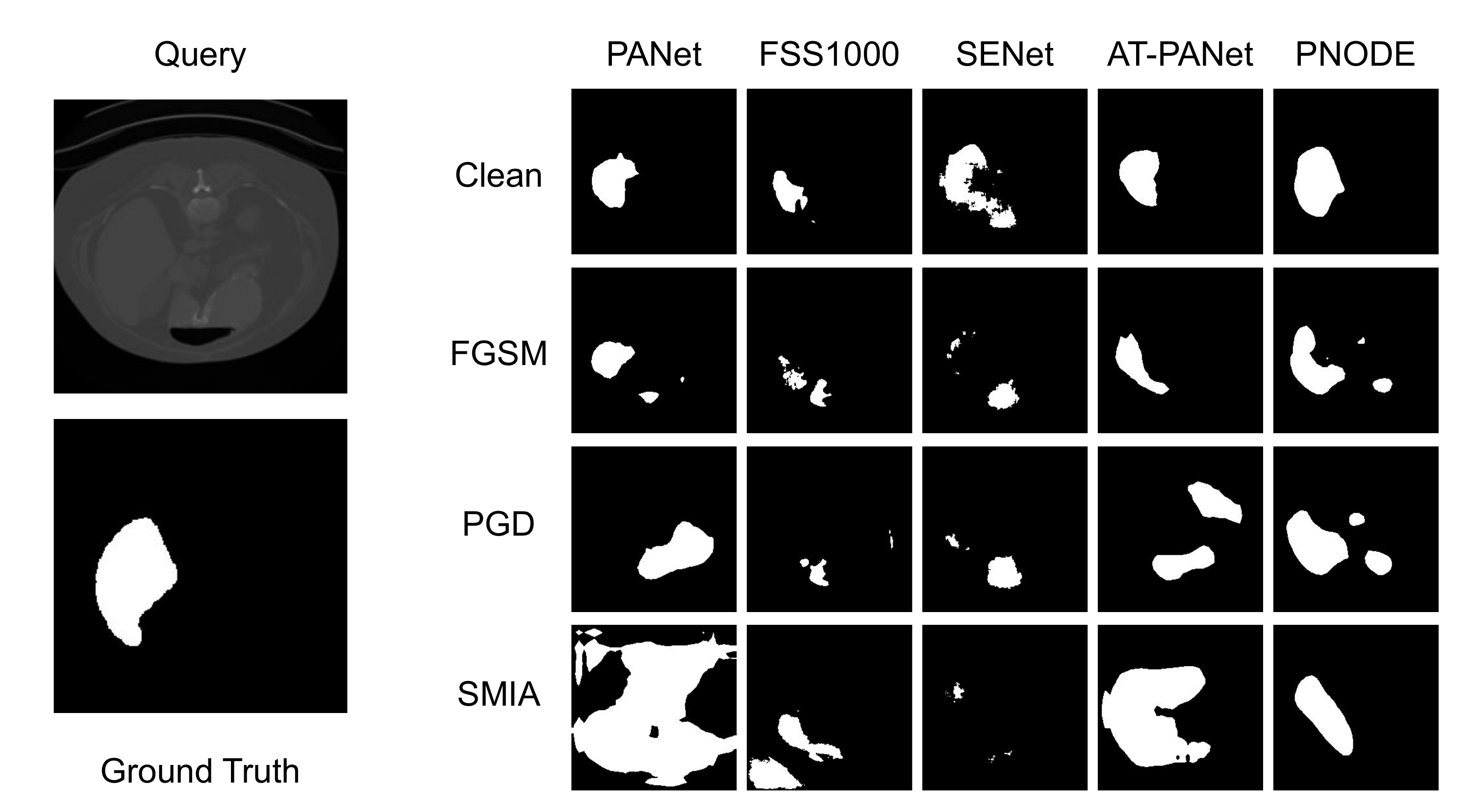}
    \caption{Predicted masks by different models for different attacks. On the left are the query sample being tested and its ground-truth.}
    \label{fig:grid}
\end{figure}
the predictions by each of these models for the different attacks in Fig.\ref{fig:grid}. For the  clean  samples, PANet, AT-PANet and PNODE are visually very similar, while FSS1000 and SENet  have relatively poorer performance.  For FGSM,  AT-PANet performs much better than PANet, most likely because of encountering similar data during training. For PGD, all of PANet, FSS1000, SENet and AT-PANet predict shapes that resemble the true organ, but are at wrong locations. PNODE is able to localise the organ better. For SMIA, all the predictions are poor and PNODE is the only one even closely  resembling the actual labels. For additional results, please refer to the supplementary material.

\section{Conclusion}

Defense against adversarial attacks on few-shot segmentation models is of utmost importance as these models are data-scarce. With their applications in the medical domain, it is critical to provide robust mechanisms against attacks of varying intensity and design. Although adversarial training can alleviate the risk associated with these attacks, the training procedure's computational overhead and poor generalizability render it less favourable as a robust defense strategy. We overcome its limitations by employing Neural-ODEs to propose adversarially robust prototypical few-shot segmentation framework PNODE that stabilizes the model against adversarial support and query attacks. PNODE is shown to have better generalization abilities against attacks while maintaining performance on clean examples. To the best of our knowledge, we are the first to study effects of different adversarial attacks on few-shot segmentation models and provide a robust defense strategy that we hope will help the medical community.

\appendix

\section{Additional Results}

\vspace{-5mm}

\begin{table}[!ht]
    \centering
    \setlength{\tabcolsep}{2.8pt}
    \renewcommand{\arraystretch}{1.2}
    \caption{1-shot \textit{support} attack results for BCV $\rightarrow$ BCV in-domain Liver and Spleen organs (novel classes). The dice scores are rounded off to two decimals.}
    \vspace{1.5mm}
    \scalebox{0.9}{
    \begin{tabular}{c|cccc|cccc}
    \toprule
        \multirow{2}{*}{Method} & \multicolumn{4}{c|}{BCV $\rightarrow$ BCV (Liver)} & \multicolumn{4}{c}{BCV $\rightarrow$ BCV (Spleen)}\\ 
    % \midrule
         & Clean & FGSM & PGD & SMIA & Clean & FGSM & PGD & SMIA \\
    \midrule
        PANet & .61±01 & .17±.06 & .39±.01& .11±.01 & .39±.05 & .17±.02 & .10±.02 & .03±.01   \\
        % FSS1000\cite{FSS} & .37±.04 & .10±.03 & .04±.02& .18±.01 & .41±.02 & .19±.01 & .08±.01 & .17±.01   \\
        % SENet\cite{SENet} & .61±.01 & .30±.06 & .22±.02& .12±.02& .57±.01 & .04±.01 & .21±.04& .01±.01 \\
      %\midrule
        AT-PANet & .65±.02 & .19±.07 & .30±.15 & .11±.01 & .49±.05 & .42±.03 & .32±.06 & .03±.01   \\
        % AT-PNODE & .62 (.00) & \textbf{.38 (.00)} & \textbf{.13 (.00)}& \textbf{.24 (.00)} & .00 (.00) & .00 (.00) & .00 (.00) & .00 (.00)   
        %\\
        \textbf{PNODE} & \textbf{.82±.03} & \textbf{.63±.17} & \textbf{.72±.02}& \textbf{.17±.05} & \textbf{.60±.02} & \textbf{.43±.01} & \textbf{.38±.02} & \textbf{.05±.01}\\
    \bottomrule
    
    \end{tabular}}
    \vspace{-10mm}
    \label{tab:bcv_1shot}
\end{table}
\begin{table}[!ht]
    \centering
    \setlength{\tabcolsep}{2.8pt}
    \renewcommand{\arraystretch}{1.2}
    \caption{1-shot \textit{support} attack results for Liver BCV $\rightarrow$ Decathlon and BCV $\rightarrow$ CT-ORG cross-domain settings. The dice scores are rounded off to two decimals.}
    \vspace{1.5mm}
    \scalebox{0.90}{
    \begin{tabular}{c|cccc|cccc}
    \toprule
        \multirow{2}{*}{Method} & \multicolumn{4}{c|}{BCV $\rightarrow$ Decathlon (Liver)} & \multicolumn{4}{c}{BCV $\rightarrow$ CT-ORG (Liver)}\\ 
    % \midrule
         & Clean & FGSM & PGD & SMIA & Clean & FGSM & PGD & SMIA \\
    \midrule
        PANet & .53±.01 & .15±.01 & .35±.02& .13±.01 & .52±.01 & .20±.05 & .43±.02 & .14±.01   \\
        % FSS1000\cite{FSS} & .49±.04 & .15±.01 & .05±.01& .19±.03 & .14±.02 & .03±.01 & .01±.01 & .08±.01   \\
        % SENet\cite{SENet} & .61 (.01) & .30 (.06) & .22 (.02)& .12 (.02)& .57 (.00) & .04 (.01) & .21 (.04)& .01 (.01) \\
      %\midrule
        AT-PANet & .57±.01 & .14±.04 & .31±.06& .13±.01 & .56±.01 & .12±.02 & .32±.08 & .14±.01   \\
        % AT-PNODE & .62 (.00) & \textbf{.38 (.00)} & \textbf{.13 (.00)}& \textbf{.24 (.00)} & .00 (.00) & .00 (.00) & .00 (.00) & .00 (.00)   
        %\\
        \textbf{PNODE} & \textbf{.65±.01} & \textbf{.49±.01} & \textbf{.55±.01}& \textbf{.16±.02} & \textbf{.66±.01} & \textbf{.51±.03} & \textbf{.50±.04} & \textbf{.22±.04}\\
    \bottomrule
    
    \end{tabular}}
    \vspace{1mm}
    \vspace{-5mm}
    \label{tab:3shot}
\end{table}

\begin{figure}
    \centering

    \includegraphics[width=0.64\linewidth]{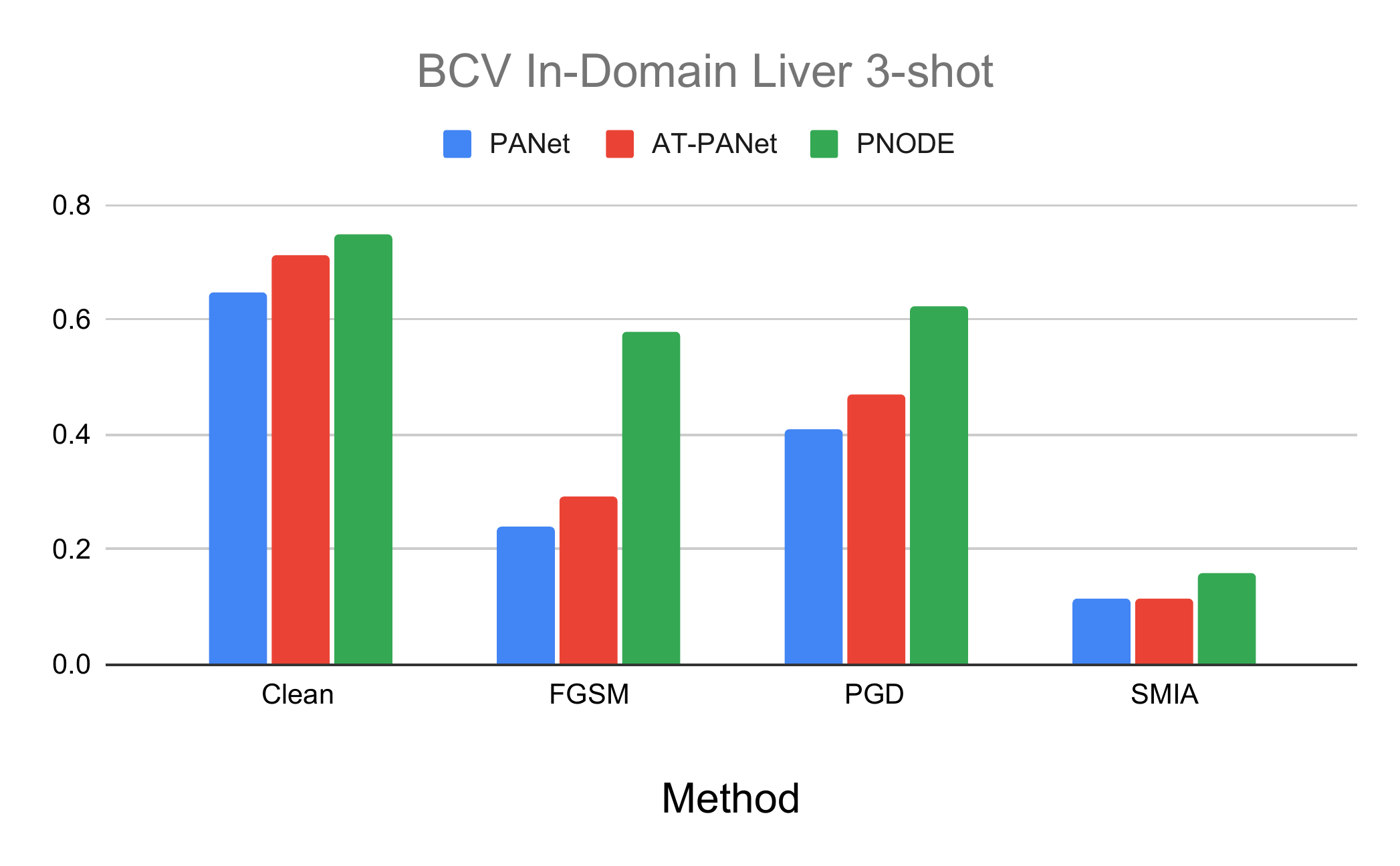}
    \includegraphics[width=0.64\linewidth]{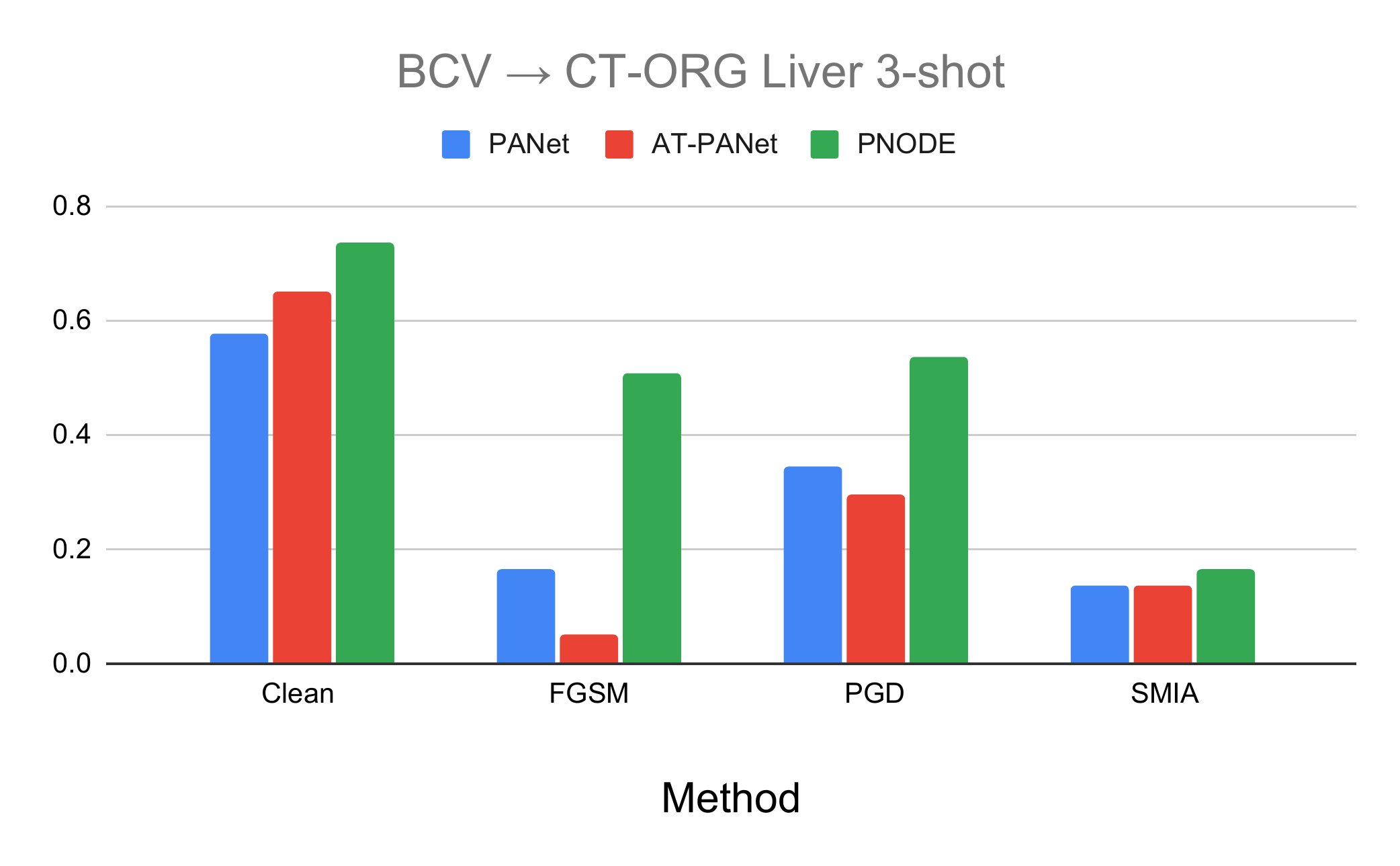}
    \vspace{-2mm}
    \caption{Performance of models for \textit{support} attacks on BCV $\rightarrow$ BCV in-domain and BCV $\rightarrow$ CT-ORG  cross-domain Liver 3-shot.}
    \label{fig:tsne}
\end{figure}

\section{Additional Plots and Visusalisations}

\vspace{0mm}
\begin{figure}
    \centering

    \includegraphics[width=0.48\linewidth]{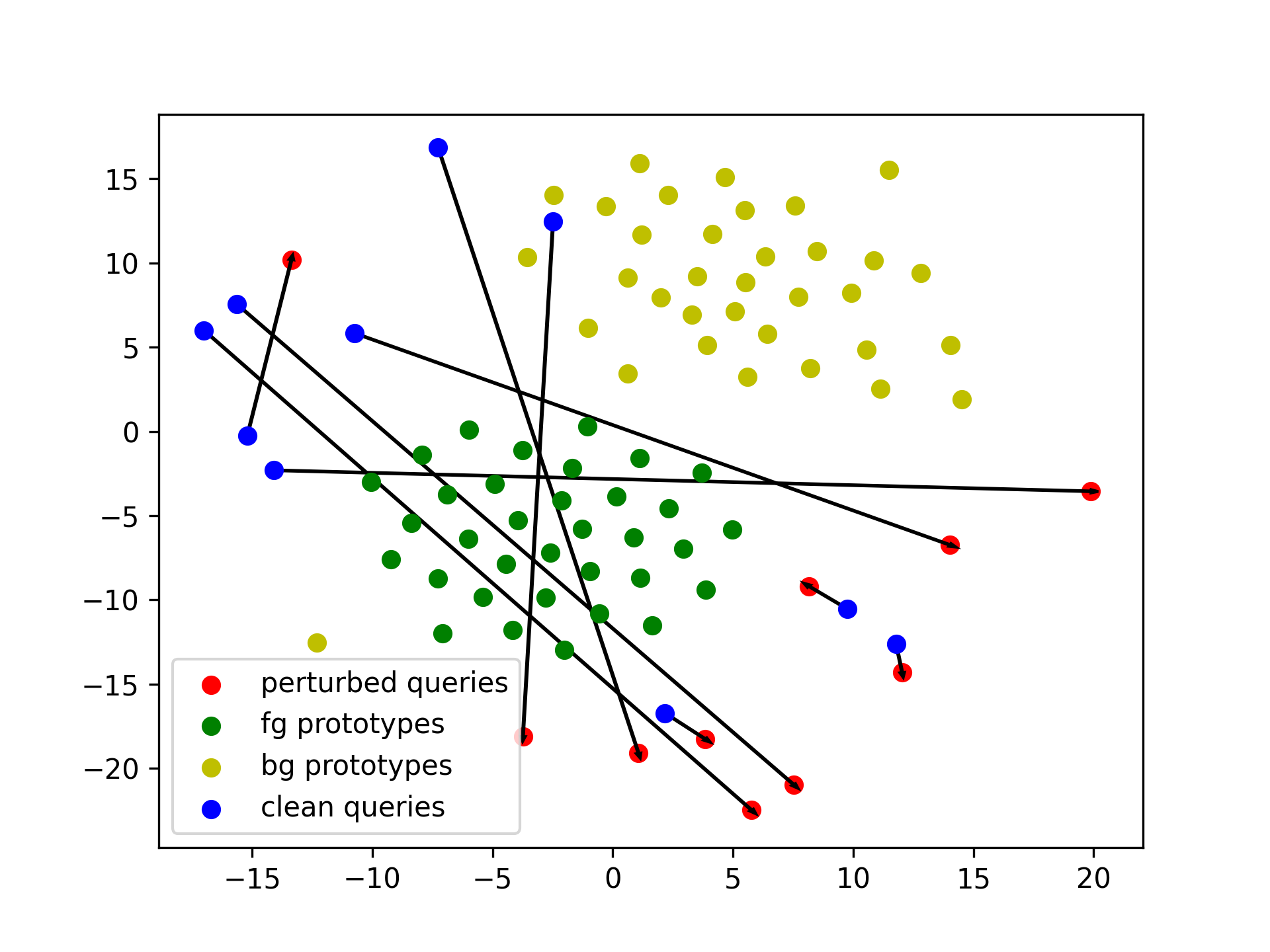}
    \includegraphics[width=0.48\linewidth]{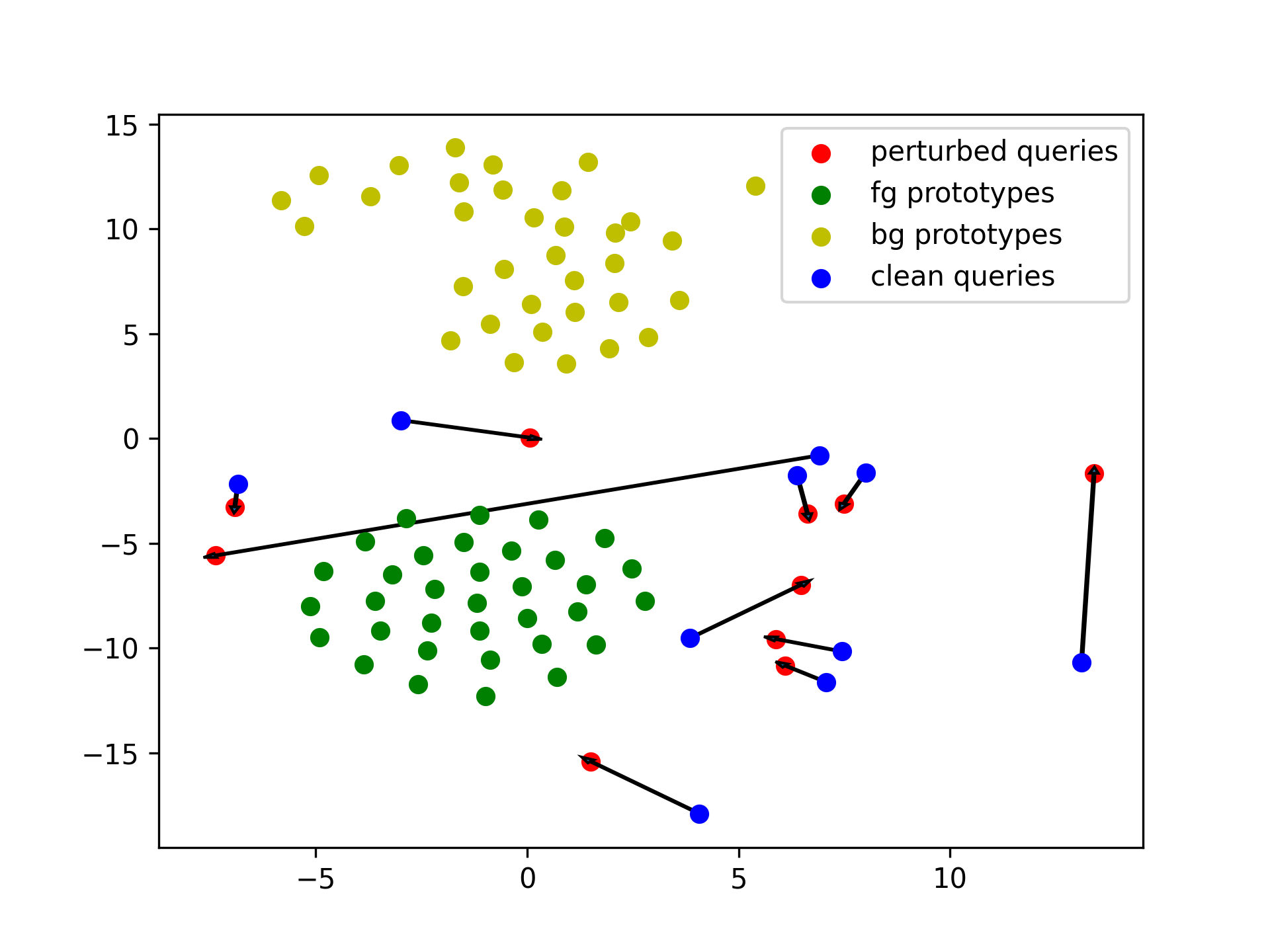}
    \vspace{-2mm}
    \caption{t-SNE visualisations of clean and perturbed queries with foreground and background prototypes for PANet (left) and PNODE (right) (BCV Liver 1-shot).}
    \label{fig:tsne}
\end{figure}
\begin{figure}
    \centering
    \vspace{-4mm}
    \includegraphics[width=0.8\linewidth]{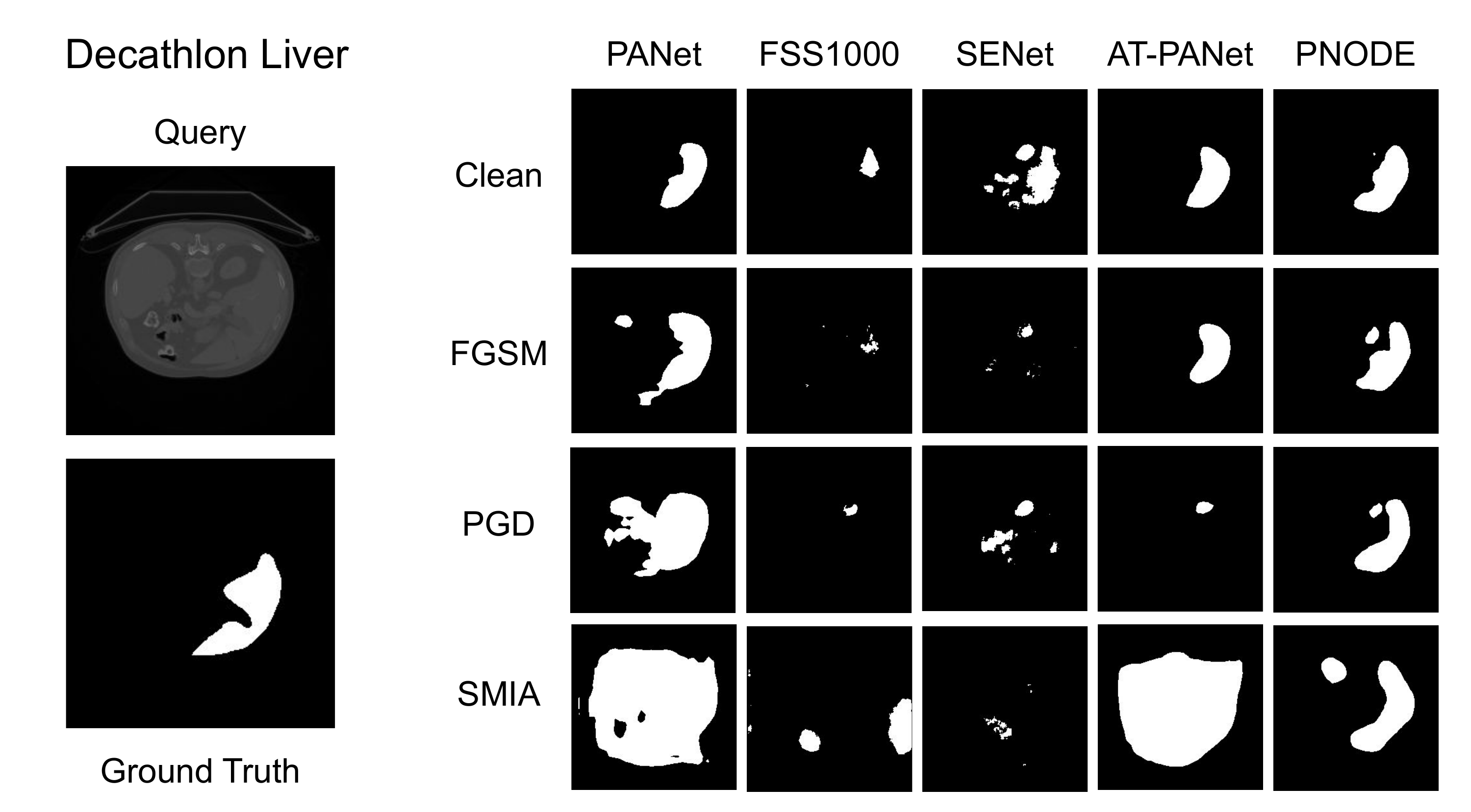}
    
    \vspace{-2mm}
    \caption{Predicted masks by different models for different attacks on Decathlon Liver 1-shot. On the left are the query sample being tested and its ground-truth.}
    \label{fig:grid_deca}
\end{figure}
\begin{figure}
    \centering
    \vspace{-4mm}
        \includegraphics[width=0.8\linewidth]{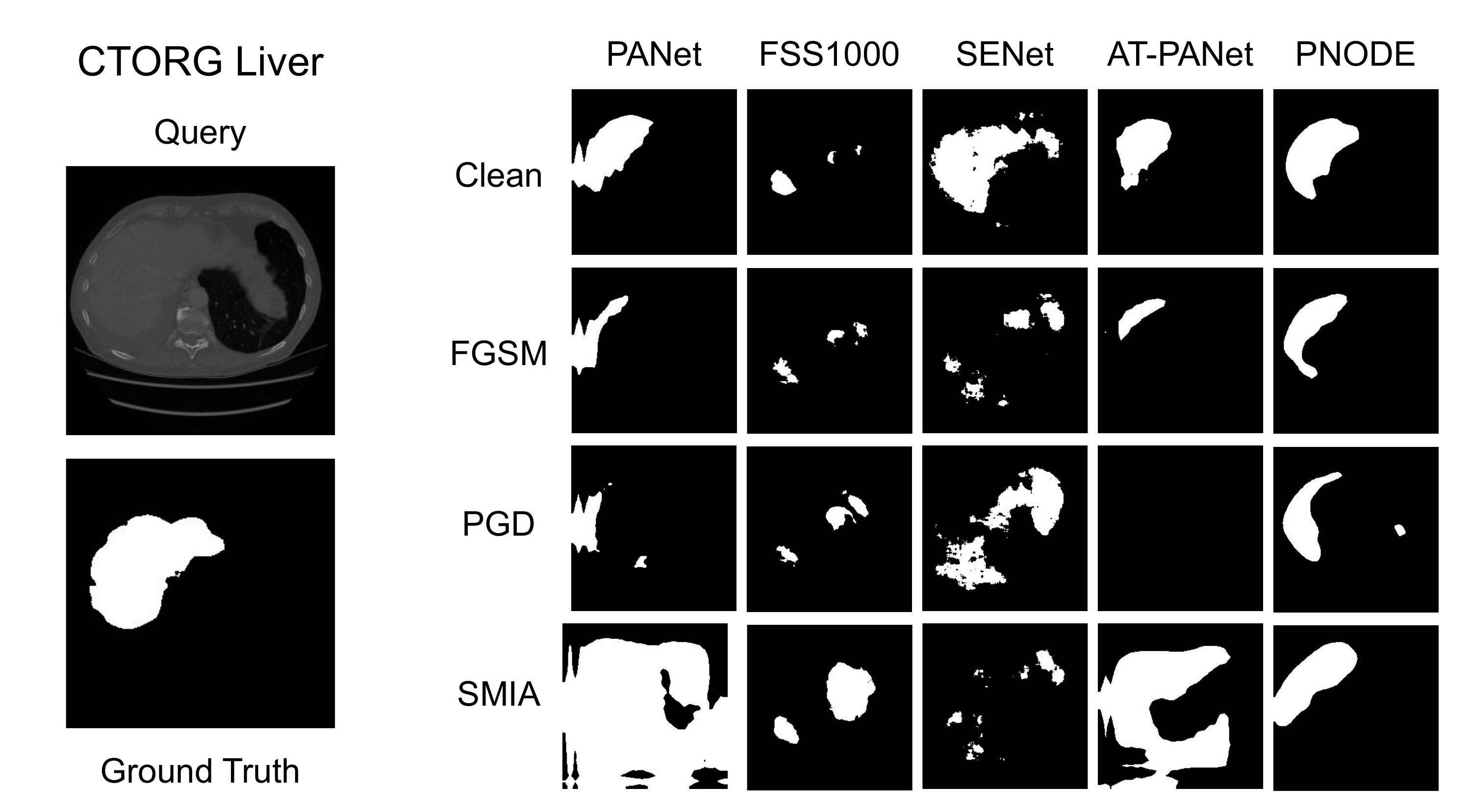}
    \vspace{-2mm}
    \caption{Predicted masks by different models for different attacks on CT-ORG Liver 1-shot. On the left are the query sample being tested and its ground-truth.}
    \label{fig:grid_ctorg}
\end{figure}


\begin{thebibliography}{8}
\bibitem{One-shot}
Li Fei-Fei, Rob Fergus, Pietro Perona. One-shot learning of object categories, IEEE TPAMI, vol. 28, 2006.
\bibitem{Ian}
Ian J. Goodfellow, Jonathon Shlens, Christian Szegedy.
Explaining and harnessing adversarial examples. In ICLR,
2015.
\bibitem{BCV}
Bennett Landman, Zhoubing Xu, Juan Eugenio Igelsias, Martin Styner, Thomas Robin. Langerak, Arno Klein. Miccai multi-atlas labeling beyond the cranial vault–workshop and challenge. In MICCAI Multi-Atlas Labeling Beyond Cranial Vault—Workshop Challenge, 2015.
\bibitem{resnet}
Kaiming He, Xiangyu Zhang, Shaoqing Ren, Jian Sun. Deep Residual Learning for Image Recognition. In CVPR, 2016.
\bibitem{Kurakin}
Alexey Kurakin, Ian J. Goodfellow, Samy Bengio. Adversarial examples in the physical world. ICLR (Workshop) 2017.
\bibitem{snell}
Jake Snell, Kevin Swersky, Richard S. Zemel. Prototypical networks for few-shot learning. In NeurIPS, 2017.
\bibitem{Ravi}
Sachin Ravi, Hugo Larochelle. Optimization as a model for few-shot learning. In ICLR, 2017.

\bibitem{Alexey}
Alexey Kurakin, Ian J. Goodfellow, Samy Bengio. Adversarial machine learning at scale. In ICLR, 2017.

\bibitem{Moosavi}
Seyed-Mohsen Moosavi-Dezfooli, Alhussein Fawzi, Omar Fawzi, Pascal Frossard. Universal
adversarial perturbations. In CVPR, 2017.

\bibitem{Xie}
Cihang Xie, Jianyu Wang, Zhishuai Zhang, Yuyin Zhou, Lingxi Xie, Alan Yuille. Adversarial
examples for semantic segmentation and object detection. In ICCV, 2017.

\bibitem{Dong}
Nanqing Dong, Eric P. Xing. Few-shot semantic segmentation
with prototype learning. In BMVC, 2018.
\bibitem{relationNet}
Flood Sung, Yongxin Yang, Li Zhang, Tao Xiang, Philip H.S. Torr, Timothy M. Hospedales. Learning to compare: Relation network for few-shot learning. In CVPR, 2018.
\bibitem{Madry}
Aleksander Madry, Aleksandar Makelov, Ludwig Schmidt, Dimitris Tsipras, Adrian Vladu. Towards deep learning models resistant to adversarial attacks. In ICLR, 2018.
\bibitem{chen2018neuralode}
Ricky T. Q. Chen, Yulia Rubanova, Jesse Bettencourt, David Duvenaud. Neural ordinary differential equations. In NeurIPS, 2018.
\bibitem{Paschali}
Magdalini Paschali, Sailesh Conjeti, Fernando Navarro, Nassir Navab.
Generalizability vs. Robustness: Investigating medical imaging networks using adversarial examples. In MICCAI, 2018.
\bibitem{Anurag}
Anurag Arnab, Ondrej Miksik, Philip H.S. Torr. On the
robustness of semantic segmentation models to adversarial
attacks. In CVPR, 2018.
\bibitem{Hongyang}
Hongyang Zhang, Yaodong Yu, Jiantao Jiao, Eric Xing, Laurent El Ghaoui, Michael Jordan. Theoretically principled trade-off between robustness and accuracy. In ICML, 2019.
\bibitem{blind}
Huan Zhang, Hongge Chen, Zhao Song, Duane Boning, Inderjit Dhillon, Cho-Jui Hsieh. The limitations of adversarial training and the blind-spot attack. In ICLR, 2019.
\bibitem{PANet}
Kaixin Wang, Jun Hao Liew, Yingtian Zou, Daquan Zhou, Jiashi Feng. Panet: Few-shot image semantic segmentation
with prototype alignment. In ICCV, 2019.

\bibitem{Zhao}
Amy Zhao, Guha Balakrishnan, Frédo Durand, John V. Guttag, Adrian V. Dalca. Data augmentation using learned
transformations for one-shot medical image segmentation. In CVPR, 2019.

\bibitem{Cheng}
Cheng Ouyang, Konstantinos Kamnitsas, Carlo Biffi, Jinming
Duan, Daniel Rueckert. Data efficient unsupervised
domain adaptation for cross-modality image segmentation.
In MICCAI, 2019.

\bibitem{Utku}
Utku Ozbulak, Arnout Van Messem, Wesley De Neve. Impact of adversarial examples on deep learning models for biomedical image segmentation. In MICCAI, 2019.

\bibitem{DECATHLON}
Amber L. Simpson, et al. 2019. A large annotated
medical image dataset for the development and evaluation of
segmentation algorithms. arXiv preprint arXiv:1902.09063.
%\bibitem{Kolter}
%J Zico Kolter and Eric Wong. Provable defenses against adversarial examples via the convex
%outer adversarial polytope. arXiv preprint arXiv:1711.00851, 2017.

%\bibitem{Wong}
%EricWong, Frank Schmidt, Jan Hendrik Metzen, and J Zico Kolter. Scaling provable adversarial
%defenses. arXiv preprint arXiv:1805.12514, 2018.
\bibitem{SENet}
Abhijit Guha Roy, Shayan Siddiqui, Sebastian Pölsterl, Nassir Navab, Christian Wachinger.
`Squeeze \& Excite' Guided few-shot segmentation of volumetric images. In MedIA, vol. 59, 2020.

\bibitem{CTORG}
Blaine Rister, Darvin Yi, Kaushik Shivakumar, Tomomi Nobashi, Daniel L. Rubin. CT-ORG, a new dataset for multiple organ segmentation in computed tomography. In Scientific Data, 2020. https://doi.org/10.1038/s41597-020-00715-8.

\bibitem{FSS}
Xiang Li, Tianhan Wei, Yau Pun Chen, Yu-Wing Tai, Chi-Keung Tang. FSS-1000: A 1000-class dataset for few-shot segmentation. In CVPR, 2020.

\bibitem{Yan}
Hanshu Yan, Jiawei Du, Vincent Y. F. Tan, Jiashi Feng. On robustness of neural ordinary differential equations. In ICLR, 2020.

\bibitem{Liu}
Xuanqing Liu, Tesi Xiao, Si Si, Qin Cao, Sanjiv Kumar, Cho-Jui Hsieh. How does noise help robustness? Explanation
and exploration under the neural sde framework. In CVPR, 2020.
\bibitem{Goldblum}
Micah Goldblum, Liam Fowl, Tom Goldstein. Adversarially robust few-shot learning: A meta-learning approach. In NeurIPS, 2020.
\bibitem{Park}
Sanglee Park, Jungmin So. On the effectiveness of adversarial training in defending against adversarial example attacks for image classification. In Applied Sciences, 10(22):8079, 2020. https://doi.org/10.3390/app10228079.

\bibitem{Kang}
Qiyu Kang, Yang Song, Qinxu Ding, Wee Peng Tay. Stable neural ode with Lyapunov-stable equilibrium points for defending against adversarial attacks. In NeurIPS, 2021.

%\bibitem{Milletari}
%Fausto Milletari, Nassir Navab, and Seyed-Ahmad Ahmadi.
%V-net: Fully convolutional neural networks for volumetric
%medical image segmentation. In 2016 Fourth International
%Conference on 3D Vision (3DV), pages 565–571.
%IEEE, 2016.
% \bibitem{Shaban}
% Amirreza Shaban, Shray Bansal, Zhen Liu, Irfan Essa, and
% Byron Boots. One-shot learning for semantic segmentation.
% arXiv preprint arXiv:1709.03410, 2017.

\bibitem{Tang}
Hao Tang, Xingwei Liu, Shanlin Sun, Xiangyi Yan, Xiaohui Xie. Recurrent mask refinement for few-shot medical image segmentation. In ICCV, 2021.

%\bibitem{jia}
%Shuai Jia, Chao Ma, Yibing Song, and Xiaokang Yang. Robust tracking %against adversarial attacks.
%In European Conference on Computer Vision, pp. 69–84. Springer, 2020.

% \bibitem{Fischer}
% Volker Fischer, Mummadi Chaithanya Kumar, Jan Hendrik Metzen, Thomas Brox. Adversarial
% examples for semantic image segmentation. arXiv preprint arXiv:1703.01101, 2017.
%\bibitem{Dong2}
%Yinpeng Dong, Tianyu Pang, Hang Su, and Jun Zhu. Evading defenses to transferable adversarial
%examples by translation-invariant attacks. In IEEE Conference on Computer Vision and Pattern
%Recognition, 2019.

%\bibitem{Finlayson}
%Samuel G Finlayson, John D Bowers, Joichi Ito, Jonathan L Zittrain, Andrew L Beam, and Isaac S
%Kohane. Adversarial attacks on medical machine learning. Science, 363(6433):1287–1289, 2019.

%\bibitem{Inception}
%Szegedy, C., Vanhoucke, V., Ioffe, S., Shlens, J., Wojna, Z.: Rethinking the inception architecture for computer vision. In: CVPR (2016)

%\bibitem{Unet}
%Ronneberger, O., Fischer, P., Brox, T.: U-Net: Convolutional Networks for Biomedical Image
%Segmentation. International Conference on Medical Image Computing and Computer-Assisted
%Intervention (2015)

%\bibitem{Haber}
%E. Haber and L. Ruthotto, “Stable architectures for deep neural networks,” Inverse Problems,
%vol. 34, no. 1, pp. 1–23, Dec. 2017.

\bibitem{Gege}
Gege Qi, Lijun Gong, Yibing Song, Kai Ma, Yefeng Zheng. Stabilized medical image attacks. In ICLR, 2021.



%\bibitem{Good}
%Ian J Goodfellow, Jonathon Shlens, and Christian Szegedy. Explaining and harnessing adversarial examples. In ICLR, 2014.



\bibitem{Divide}
Xiaogang Xu, Hengshuang Zhao, Jiaya Jia. Dynamic divide-and-conquer adversarial training for robust semantic segmentation. In ICCV, 2021.


% \bibitem{VGG}
% Simonyan, K. and Zisserman, A., 2014. Very deep convolutional networks for large-scale image recognition. arXiv preprint arXiv:1409.1556.


% \bibitem{Imagenet}
% J. Deng, W. Dong, R. Socher, L. -J. Li, Kai Li and Li Fei-Fei, "ImageNet: A large-scale hierarchical image database," 2009 IEEE Conference on Computer Vision and Pattern Recognition, 2009.

%\bibitem{torchdiffeq}
%https://github.com/rtqichen/torchdiffeq


\end{thebibliography}
\end{document}